\documentclass{article}

\usepackage{microtype}
\usepackage{graphicx}
\usepackage{subfigure}
\usepackage{booktabs} 

\usepackage{hyperref}

\newcommand{\longmodel}{Bayesian Inverse Planning and Core Knowledge}
\newcommand{\model}{BIPaCK}

\usepackage[accepted]{icml2021}

\usepackage{dsfont}
\usepackage{amsmath,amssymb}
\DeclareMathOperator*{\argmax}{arg\,max}
\DeclareMathOperator{\E}{\mathbb{E}}

\usepackage{array}
\newcolumntype{L}[1]{>{\raggedright\let\newline\\\arraybackslash\hspace{0pt}}m{#1}}
\newcolumntype{C}[1]{>{\centering\let\newline\\\arraybackslash\hspace{0pt}}m{#1}}
\newcolumntype{R}[1]{>{\raggedleft\let\newline\\\arraybackslash\hspace{0pt}}m{#1}}

\usepackage{multirow}

\icmltitlerunning{AGENT: A Benchmark for Core Psychological Reasoning}

\begin{document}

\twocolumn[
\icmltitle{AGENT: A Benchmark for Core Psychological Reasoning}



\icmlsetsymbol{equal}{*}

\begin{icmlauthorlist}
\icmlauthor{Tianmin Shu}{mit}
\icmlauthor{Abhishek Bhandwaldar}{ibm}
\icmlauthor{Chuang Gan}{ibm}
\icmlauthor{Kevin A. Smith}{mit}
\icmlauthor{Shari Liu}{mit}
\icmlauthor{Dan Gutfreund}{ibm}
\icmlauthor{Elizabeth Spelke}{harvard}
\icmlauthor{Joshua B. Tenenbaum}{mit}
\icmlauthor{Tomer D. Ullman}{harvard}
\end{icmlauthorlist}

\icmlaffiliation{mit}{Massachusetts Institute of Technology}
\icmlaffiliation{ibm}{MIT-IBM Watson AI Lab}
\icmlaffiliation{harvard}{Harvard University}

\icmlcorrespondingauthor{Tianmin Shu}{tshu@mit.edu}

\icmlkeywords{Machine Learning, ICML}

\vskip 0.3in
]



\printAffiliationsAndNotice{}  

\begin{abstract}
For machine agents to successfully interact with humans in real-world settings, they will need to develop an understanding of human mental life. Intuitive psychology, the ability to reason about hidden mental variables that drive observable actions, comes naturally to people: even pre-verbal infants can tell agents from objects, expecting agents to act efficiently to achieve goals given constraints. Despite recent interest in machine agents that reason about other agents, it is not clear if such agents learn or hold the core psychology principles that drive human reasoning. Inspired by cognitive development studies on intuitive psychology, we present a benchmark consisting of a large dataset of procedurally generated 3D animations, AGENT (Action, Goal, Efficiency, coNstraint, uTility), structured around four scenarios (goal preferences, action efficiency, unobserved constraints, and cost-reward trade-offs) that probe key concepts of core intuitive psychology. We validate AGENT with human-ratings, propose an evaluation protocol emphasizing generalization, and compare two strong baselines built on Bayesian inverse planning and a Theory of Mind neural network. Our results suggest that to pass the designed tests of core intuitive psychology at human levels, a model must acquire or have built-in representations of how agents plan, combining utility computations and core knowledge of objects and physics.\footnote{The dataset and the supplementary material are available at \url{https://www.tshu.io/AGENT}.}
\end{abstract}

\section{Introduction}

\begin{figure}[t!]
\centering
\includegraphics[width=0.46\textwidth]{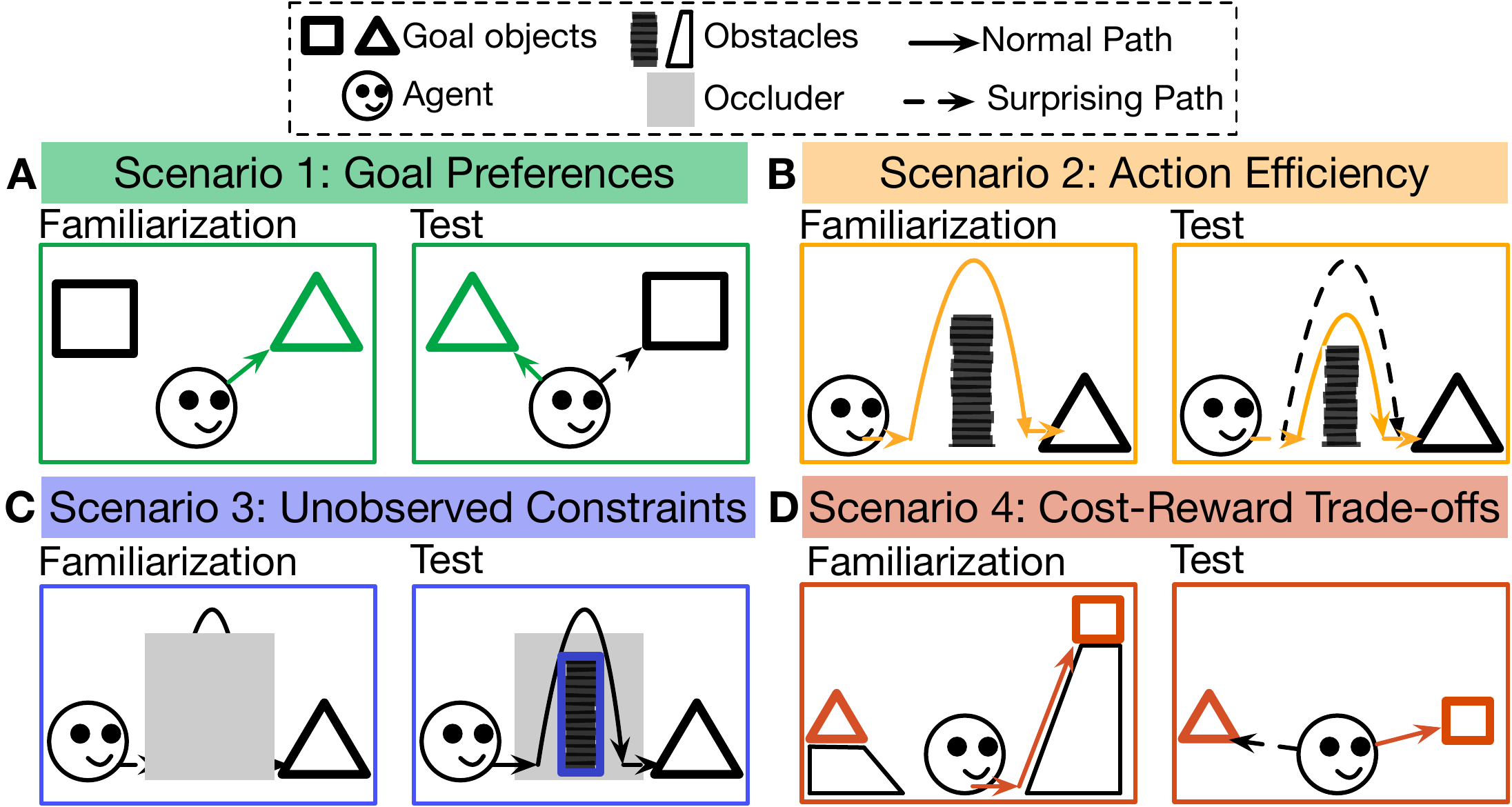}
\vspace{-5pt}
\caption{Schematic of the four key scenarios of core intuitive psychology evaluated in AGENT. Each scenario is color coded. Solid arrows show the typical behavior of the agent in the familiarization video(s) or in the expected test video. Dashed arrows show agent behavior in the surprising test video. In Unobserved Constraints trials (C), a surprising test video shows an unexpected outcome (e.g. no barrier) behind the occluder.}
\vspace{-0.5em}
\label{fig:intro}
\end{figure}

In recent years, there has been a growing interest in building socially-aware agents that can interact with humans in the real world \cite{dautenhahn2007socially,sheridan2016human,puig2020watch}. This requires agents that understand the motivations and actions of their human counterparts, an ability that comes naturally to people. Humans have an early-developing intuitive psychology, the ability to reason about other people's mental states from observed actions. From infancy, we can easily differentiate agents from objects, expecting agents to not only follow physical constraints, but also to act efficiently to achieve their goals given constraints. Even pre-verbal infants can recognize other people's costs and rewards, infer unobserved constraints given partially observed actions, and predict future actions \cite{Baillargeon2016,Gergely2003,Liu2017,Woodward1998}. This early core psychological reasoning develops with limited experience, yet generalizes to novel agents and situations, and forms the basis for commonsense psychological reasoning later in life.

Like human infants, it is critical for machine agents to develop an adequate capacity of understanding human minds, in order to successfully engage in social interactions. Recent work has demonstrated promising results towards building agents that can infer the mental states of others \cite{baker2017rational, rabinowitz2018machine}, predict people's future actions \cite{kong2018human}, and even work with human partners \cite{rozo2016learning,carroll2019utility}.  However, to date there has been a lack of rigorous evaluation benchmarks for assessing how much artificial agents learn about core psychological reasoning, and how well their learned representations generalize to novel agents and environments.

In this paper, we present AGENT (Action, Goal, Efficiency, coNstraint, uTility), a benchmark for core psychology reasoning inspired by experiments in cognitive development that probe young children's understanding of intuitive psychology. AGENT consists of a large-scale dataset of 3D animations of an agent moving under various physical constraints and interacting with various objects. These animations are organized into four categories of trials, designed to probe a machine learning model's understanding of key situations that have served to reveal infants’ intuitive psychology, testing their attributions of goal preferences (Figure~\ref{fig:intro}A; \citealt{Woodward1998}), action efficiency (Figure~\ref{fig:intro}B; \citealt{Gergely1995Efficiency}), unobserved constraints (Figure~\ref{fig:intro}C; \citealt{Csibra2003Hidden}), and cost-reward trade-offs (Figure~\ref{fig:intro}D; \citealt{Liu2017}). As we detail in Section~\ref{sec:dataset_overview}, each scenario is based on previous developmental studies, and is meant to test a combination of underlying key concepts in human core psychology. These scenarios cover the early understanding of agents as self-propelled physical entities that value some states of the world over others, and act to maximize their rewards and minimize costs subject to constraints. In addition to this minimal set of concepts, a model may also need to understand other concepts to pass a full battery of core intuitive psychology, including perceptual access and intuitive physics. Although this minimal set does not include other concepts of intuitive psychology such as false belief, it is considered part of `core psychology' in young children who cannot yet pass false belief tasks, and forms the building blocks for later concepts like false belief.


Like experiments in many infant studies, each trial has two phases: in the \textit{familiarization} phase, we show one or more videos of a particular agent's behavior in certain physical environments to a model; then in the \textit{test} phase, we show the model a video of the behavior of the same agent in a new environment, which either is `expected' or `surprising,' given the behavior of the agent in familiarization. The model's task is to judge how surprising the agent's behaviors in the test videos are, based on what the model has learned or inferred about the agent's actions, utilities, and physical constraints from watching the familiarization video(s). We validate AGENT with large-scale human-rating trials, showing that on average, adult human observers rate the `surprising' test videos as more surprising than the `expected' test videos.

   


Unlike typical evaluation for Theory of Mind reasoning \cite{rabinowitz2018machine}, we propose an evaluation protocol focusing on generalization. 
We expect models to perform well not only in test trials similar to those from training, but also in test trials that require generalization to different physical configurations within the same scenario, or to other scenarios. We compare two strong baselines for Theory of Mind reasoning: (i) \longmodel, which combines Bayesian inverse planning \cite{baker2017rational} with physical simulation \cite{battaglia2013simulation}, and (ii) ToMnet-G, which extends the Theory of Mind neural network \cite{rabinowitz2018machine}. Our experimental results show that ToMnet-G can achieve reasonably high accuracy when trained and tested on trials of similar configurations or of the same scenario, but faces a strong challenge of generalizing to different physical situations, or a different but related scenario. In contrast, due to built-in representations of planning, objects, and physics, \model~achieves a stronger performance on generalization both within and across scenarios. This demonstrates that AGENT poses a useful challenge for building models that achieve core psychological reasoning via learned or built-in representations of agent behaviors that integrate utility computations, object representations, and intuitive physics.

In summary, our contributions are: (i) a new benchmark on core psychological reasoning consisting of a large-scale dataset inspired by infant cognition and validated by human trials, (ii) a comprehensive comparison of two strong baseline models that extends prior approaches for mental state reasoning, and (iii) a generalization-focused evaluation protocol. We plan to release the dataset and the code for data generation.


\section{Related Work}

\noindent\textbf{Machine Social Perception.} While there has been a long and rich history in machine learning concerning human behavior recognition \cite{aggarwal2011human,caba2015activitynet,poppe2010survey,choi2013understanding,shu2015joint,ibrahim2016hierarchical,sigurdsson2018charades,fouhey2018lifestyle} and forecasting \cite{kitani2012activity,koppula2013learning,alahi2016social,kong2018human,liang2019peeking}, prior work has typically focused on classifying and/or predicting motion patterns. However, the kind of core psychological reasoning evaluated in AGENT emphasizes mental state reasoning. This objective is loosely aligned with agent modeling in work on multi-agent cooperation or competition \cite{albrecht2018autonomous}, where a machine agent attempts to model another agent's type, defined by factors such as intentions \cite{mordatch2018emergence,puig2020watch}, rewards \cite{abbeel2004apprenticeship, ziebart2008maximum, hadfield2016cooperative, shu2018m}, or policies \cite{sadigh2016planning,kleiman2016coordinate,nikolaidis2017human,lowe2017multi,wang2020too,xie2020learning}. In addition, the recent interest in value alignment \cite{hadfield2016cooperative} is also essentially about learning key aspects of intuitive psychology, including goal preferences, rewards, and costs. Here, we present a rigorously designed and human-validated dataset for benchmarking a machine agent's ability to model aspects of other agents' mental states that are core to human intuitive psychology. These protocols can be used in future work to build and test models that reason and learn about other minds the way that humans do.

\noindent\textbf{Synthetic Datasets for Machine Perception.} Empowered by graphics and physics simulation engines, there have been synthetic datasets for various problems in machine scene understanding \cite{zitnick2014adopting,ros2016synthia,johnson2017clevr,song2017semantic,xiazamirhe2018gibsonenv,riochet2018intphys,jiang2018configurable,groth2018shapestacks,crosby2019animal,yi2019clevrer,bakhtin2019phyre,nan2020learning,phase}. Many of these datasets focusing on social perception are either built using simple 2D cartoons \cite{zitnick2014adopting,gordon2016commonsense,phase}, or focus on simpler reasoning tasks \cite{cao2020long}. Concurrent with this paper, \citealt{Gandhi2021Baby} have proposed a benchmark, BIB (Baby Intuitions Benchmark), for probing a model's understanding of other agents' goals, preferences, actions in maze-like environments. The tests proposed in AGENT have conceptual overlap with BIB, with three key differences: First, in addition to the common concepts tested in both benchmarks (goals, preferences, and actions), the scenarios in AGENT probe concepts such as unobserved constraints and cost-reward trade-offs, whereas BIB focuses on the instrumentality of actions (e.g., using a sequence of actions to make an object reachable before getting it). Second, trials in AGENT simulate diverse physical situations, including ramps, platforms, doors, and bridges, while BIB contains scenes that require more limited knowledge of physical constraints: mazes with walls. Third, the evaluation protocol for AGENT emphasizes generalization across different scenarios and types of trials, while BIB focuses on whether intuitive psychology concepts can be learned and utilized from a single large training set in the first place. BIB also provides baseline models that build on raw pixels or object masks, while our baseline models address the separate challenges presented by AGENT and focus more on incorporating the core knowledge of objects and physics into the psychological reasoning. We see that AGENT and BIB provide complementary tools for benchmarking machine agents' core psychology reasoning, and relevant models could make use of both.  

\noindent\textbf{Few-shot Imitation Learning.} The two-phase setup of the trials in AGENT resembles few-shot imitation learning \cite{duan2017one,finn2017one,yu2018one,james2018task,huang2019continuous,silver2020few}, where the objective is to imitate expert policies on multiple tasks based on a set of demonstrations. This is critically different from the objective of our benchmark, which is to asses how well models infer the mental states of a particular agent from a single or few familiarization videos, and predict the same agent's behavior in a different physical situation.


\section{AGENT Dataset}

\subsection{Overview}\label{sec:dataset_overview}

\begin{figure*}[t!]
\centering
\includegraphics[width=1.0\textwidth]{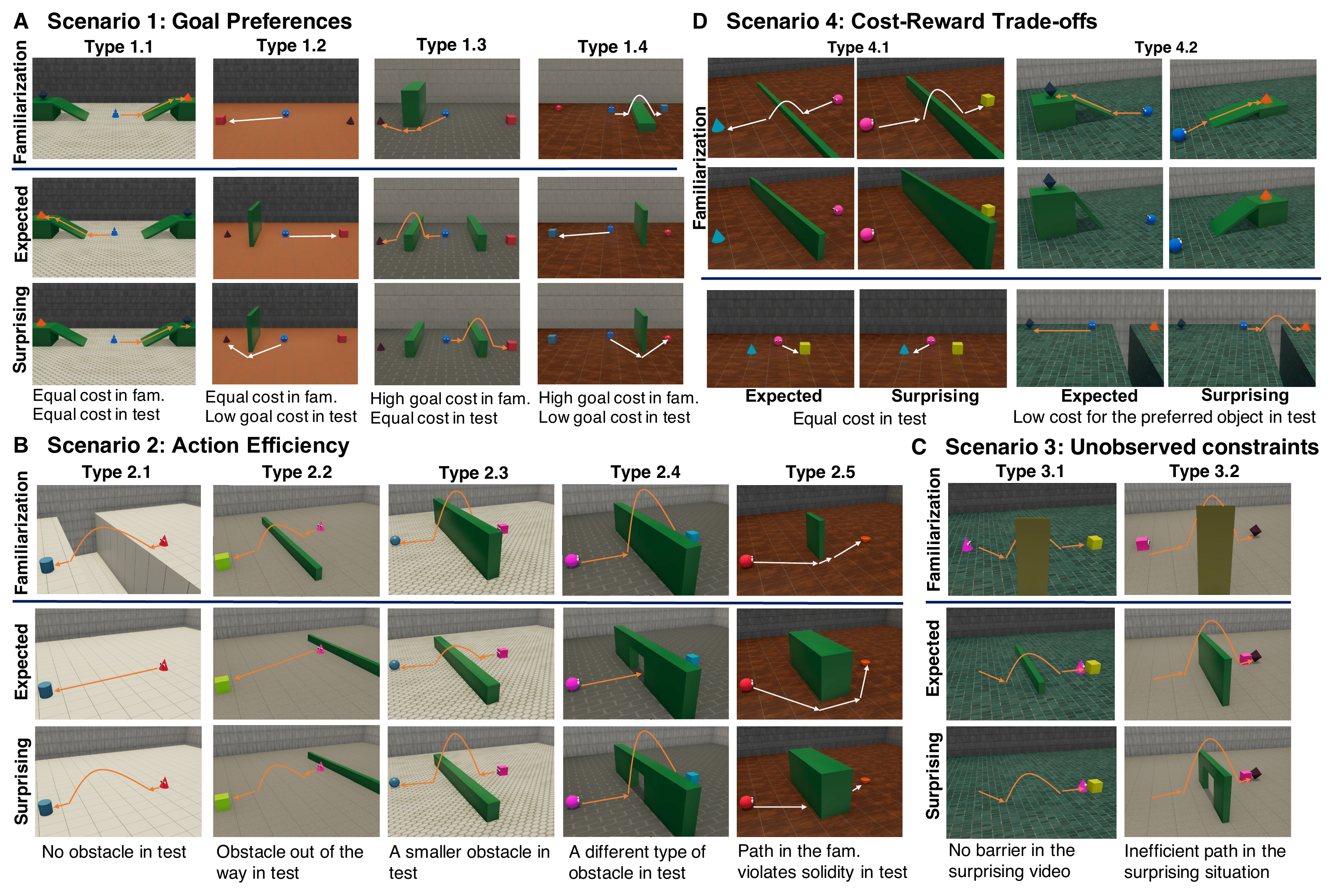}
\vspace{-20pt}
\caption{Overview of trial types of four scenarios in AGENT. Each scenario is inspired by infant cognition and meant to test a different facet of intuitive psychology. Each type controls for the possibility of learning simpler heuristics. Example videos can be viewed at \url{https://www.tshu.io/AGENT}.}
\label{fig:dataset_overview}
\end{figure*}

Figure~\ref{fig:dataset_overview} summarizes the design of trials in AGENT, which groups trials into four scenarios. All trials have two phases: (i) a familiarization phase showing one or multiple videos of the typical behaviors of a particular agent, and (ii) a test phase showing a single video of the same agent either in a new physical situation (the Goal Preference, Action Efficiency and Cost-Reward Trade-offs scenarios) or the same video as familiarization but revealing a portion of the scene that was previously occluded (Unobserved Constraints). Each test video is either \textit{expected} or \textit{surprising}. In an \textit{expected} test video, the agent behaves consistently with its actions from the familiarization video(s) (e.g. pursues the same goal, acts efficiently with respect to its constraints, and maximizes rewards), whereas in a \textit{surprising} test video, the agent aims for a goal inconsistent with its actions from the familiarization videos, achieves its goal inefficiently, or violates physics. Each scenario has several variants, including both basic versions replicating stimuli used in infant studies, and additional types with new setups of the physical scenes, creating more diverse scenarios and enabling harder tests of generalization.

\noindent\textbf{Scenario 1: Goal Preferences.} This subset of trials probes if a model understands that an agent chooses to pursue a particular goal object based on its preferences, and that pursuing the same goal could lead to different actions in new physical situations, following \citet{Woodward1998}. Each trial includes one familiarization video and a test video, where two distinct objects (with different shapes and colors) are placed on either side of an agent. For half of the test videos, the positions of the objects change from familiarization to test. During familiarization, the agent prefers one object over the other, and always goes to the preferred object. In a expected test video, the agent goes to the preferred object regardless of where it is, whereas in a surprising test video, the agent goes to the less preferred object. A good model should expect a rational agent to pursue its preferred object at test, despite the varying physical conditions. To show a variety of configurations and thus control for low level heuristics, we define four types of trials for the Goal Preferences scenario (Figure~\ref{fig:dataset_overview}), that vary the relative cost to pursue either one of the goal objects in the familiarization video and the test video. In Type 1.1 and Type 1.2, reaching either one of the objects requires the same effort as during familiarization, whereas in Type 1.3 and Type 1.4, the agent needs to overcome a harder obstacle to reach its preferred object. In Type 1.1 and Type 1.3, the agent needs to overcome the same obstacle to reach either object in the test video, but reaching the less desired object in the test video of Type 1.2 and Type 1.4 requires a higher effort for the agent than reaching the preferred object does.

\noindent\textbf{Scenario 2: Action Efficiency.} This task evaluates if a model understands that a rational agent is physically constrained by the environment and tends to take the most efficient action to reach its goal given its particular physical constraints (e.g., walls or gaps in the floor). This means that an agent may not  follow the same path for the same goal if the physical environment is no longer the same as before.
In the familiarization video, we show an agent taking an efficient path to reach a goal object given the constraints. In Type 2.1, that constraint is removed, and at test, agent takes a more efficient path (expected), or takes the same path as it had with the constraint in place (surprising). Types 2.2-4 further extend this scenario by ensuring that a model cannot use the presence of the obstacle to infer that an agent should jump by placing the obstacle out of the way (2.2), using a smaller obstacle (2.3), or introducing a door or a bridge into the obstacle (2.4). By introducing a surprising path in which the agent moves through the wall, Type 2.5 ensures that the model is not simply ignoring constraints and predicting that the closest path to a straight line is the most reasonable.

\noindent\textbf{Scenario 3: Unobserved Constraints.}
By assuming that agents tend to take the most efficient action to reach their goals (Scenarios 1-2), infants are also able to infer hidden obstacles based on agents' actions. Specifically, after seeing an agent that performs a costly action (e.g. jumps up and lands behind an occluder), infants can infer that there must be an unobserved physical constraint (e.g. a obstacle behind the occluder) that explains this action \cite{Csibra2003Hidden}. To evaluate if a model can reason about hidden constraints in this way, we designed two types of trials for Scenario 3. In both types of trials, we show an agent taking curved paths to reach a goal object (either by jumping vertically or moving horizontally), but the middle of the agent's path is hidden behind an occluder (the wall appearing in the middle of the familiarization video in Figure~\ref{fig:dataset_overview}C). In these videos, the occluder partially hides the agent from view, and it is clear that the agent is deviating from a straight path towards its goal. In the test videos, the occluder falls after the agent reaches goal object, potentially revealing the unseen physical constraints. Similar to \citet{Csibra2003Hidden}, in the expected video, the occluder falls to reveal an obstacle that justifies the action that the agent took as efficient; in the surprising video, the occluder falls to reveal an obstacle that makes the observed actions appear inefficient. 
The videos of Type 3.2 control for the absence of an object behind the occluder being a signal for surprise by revealing an obstacle that nonetheless makes the agent's actions inefficient (a smaller wall that the agent could have leapt over or moved around with less effort, or a wall with a doorway that the agent could have passed through).

\noindent\textbf{Scenario 4: Cost-Reward Trade-offs.}
Scenario 1 requires reasoning about preferences over different goal states, and Scenarios 2 and 3 require reasoning about cost functions and physical constraints. However, infants can do more than reason about agents' goals and physically grounded costs in isolation. They can also infer what goal objects agents prefer from observing the level of cost they willingly expend for their goals \cite{Liu2017}. To succeed here, infants need to understand that agents plan actions based on utility, which can be decomposed into positive rewards and negative costs \cite{Jara-Ettinger2016NUC}. Rational action under this framework thus requires agents (and observers of their actions) to trade off the rewards of goal states against the costs of reaching those goal states. Following experiments designed to probe infants' understanding of rewards and costs \cite{Liu2017}, we construct two types of trials for Scenario 4. Here we show the agent acting towards each of two goal objects under two different physical situations (four familiarization videos in total). In the first two familiarization videos, the agent overcomes an obstacle with a medium difficulty (a wall/platform/ramp with a medium height, or a chasm with a medium width) to reach the object that it likes more, but gives up when the obstacle becomes too difficult (e.g., the maximum height or width). In the remaining two familiarization videos, the agent overcomes an easy obstacle to reach the less preferred object, but decides not to pursue the same object when there is a medium-difficulty obstacle. During the testing phase, both objects are present in the scene for the first time. The agent goes to the more preferred object in the expected video, but goes to the less preferred object in the surprising video. Type 4.1 shows no obstacles, or obstacles of the same difficulty, between the agent and the two objects in the test videos. In Type 4.2, a more difficult obstacle is placed between the agent and the less preferred object at test. In both cases, a rational agent will tend to choose the object it likes more, which requires either the same amount of action cost to reach as the less preferred object (Type 4.1) or even less action cost than the less preferred object (Type 4.2). The key question is whether the model can infer this preference from the familiarization videos, and generalize it to the test video.

We introduce the human inductive biases in these four scenarios for two main reasons: (1) Human inductive biases are useful starting points for machine models, likely to help find better reward/cost functions than the ones based on raw states, and improve sample efficiency. Prior work on inverse reinforcement learning emphasizes the importance of human inductive biases for engineering useful features for the reward functions, such as the ``known features'' assumption in \cite{abbeel2004apprenticeship}. (2) Even if an AI can find a good, non-human-like reward function without human biases, a machine agent that successfully interacts with people needs to predict and reason about human intuition (Hadfield-Menell et al., 2016). In such cases, inductive biases serve as common ground to promote mutual understanding.

\begin{figure}[t!]
\centering
\includegraphics[width=0.48\textwidth]{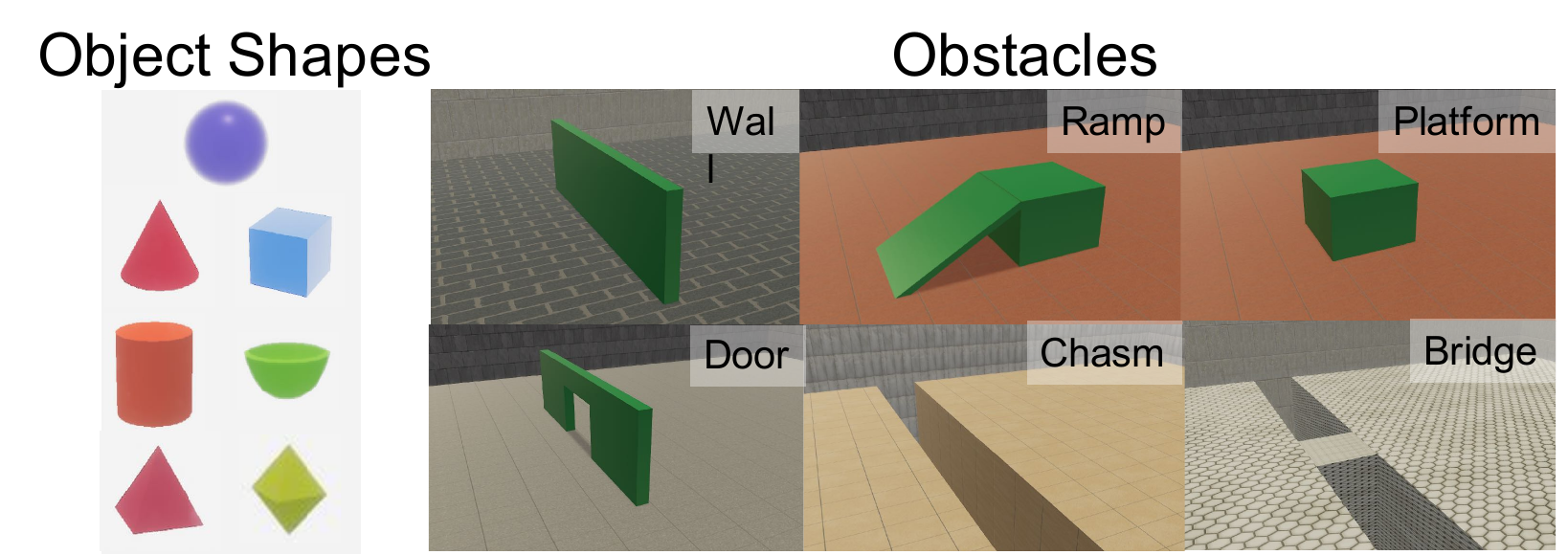}
\vspace{-10pt}
\caption{Object shapes and obstacles used in AGENT. }
\label{fig:shapes_barriers}
\vspace{-10pt}
\end{figure}

\subsection{Procedural Generation}

To generate each trial, we first sample a physical scene graph for each familiarization and test video that satisfies the constraints specified for each trial type. In this scene graph, we define the number, types, and sizes of obstacles (e.g., walls, ramps, etc.), the texture of the floor (out of 8 types), the texture of the background wall (out of 3 types), as well as the shapes, colors, sizes, and the initial positions of the agent and all objects. We then instantiate the scene graph in an open sourced 3D simulation environment, TDW \cite{gan2020threedworld}. We define the goal of the agent in each trial by randomly assign preferences of objects to the agent, and simulate the agent's path through the environment using (i) hand-crafted motion heuristics such as predefined way points and corresponding actions (i.e., walking, jumping, climbing) to reach each way point in order to overcome an obstacle of certain type and size, and (ii) a gaze turning motion that is naturally aligned with behaviors such as looking at the surrounding at beginning and looking forward while moving. We sample object shapes and obstacles from the set depicted in Figure~\ref{fig:shapes_barriers}. Note that agent shapes are always sampled from the sphere, cone, and cube subset.



\subsection{Dataset Structure}
There are 8400 videos in AGENT. Each video lasts from 5.6 s to 25.2 s, with a frame rate of 35 fps. With these videos, we constructed 3360 trials in total, divided into 1920 training trials, 480 validation trials, and 960 testing trials (or 480 pairs of expected and surprising testing trials, where each pair shares the same familiarization video(s)). All training and validation trials only contain expected test videos. 

In the dataset, we provide RGB-D frames, instance segmentation maps, and the camera parameters of the videos as well as the 3D bounding boxes of all entities recorded from the TDW simulator. We categorize entities into three classes: agent, object, and obstacle, which are also available. For creating consistent identities of the objects in a trial, we define 8 distinct colors and assign the corresponding color codes of the objects in the ground-truth information as well.

\subsection{Dataset Usage}
As our experimental results in Section~\ref{sec:exp_gen} show, training from scratch on just our dataset will not work. Instead, we suggest that to pass the tests, it is necessary to acquire additional knowledge, either via inductive biases in the architectures, or from training on additional data. Specifically, learning within this dataset may focus on extracting and representing (1) the dynamic 3D scenes, and (2) the agent properties in the familiarization trials. Additional training may follow a modular paradigm (training different model components such as perception, or planning on other datasets), or a finetuning paradigm (model components trained on other datasets could be finetuned with our training trials).

\section{Baseline Methods}

We propose two strong baseline methods for the benchmark built on well-known approaches to Theory of Mind reasoning. We provide a sketch of both methods here, and discuss implementation details in the supplementary material.

\begin{figure}[t!]
\centering
\includegraphics[width=0.48\textwidth]{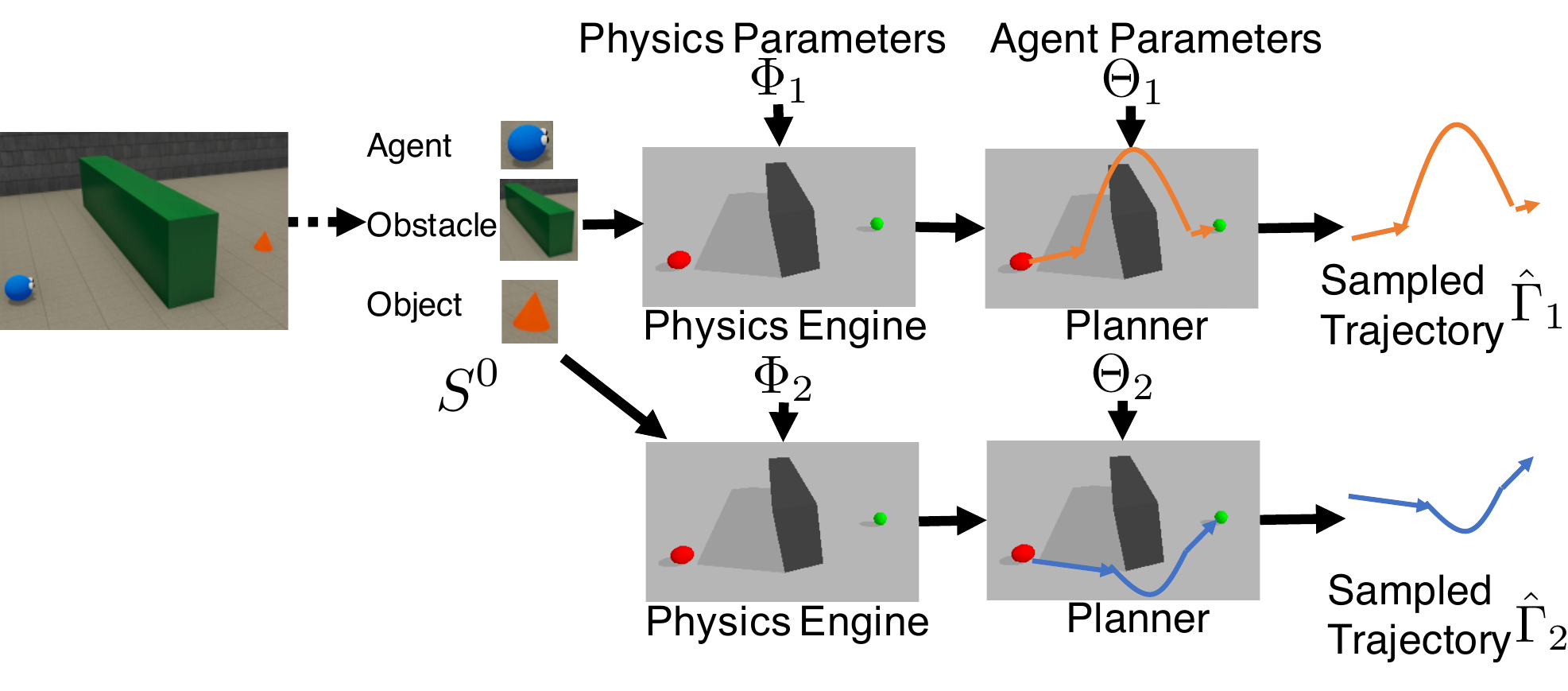}
\vspace{-15pt}
\caption{Overview of the generative model for \model. The dashed arrow indicates extracting states via the ground-truth or a perception model.}
\label{fig:InvPlan}
\vspace{-10pt}
\end{figure}

\subsection{\longmodel}\label{sec:BIP}

The core idea of Bayesian inverse planning is to infer hidden mental states (such as goals, preferences, and beliefs), through a generative model of an agent's plans \cite{baker2017rational}. Combined with core knowledge of physics \cite{baillargeon1996physical,Spelke1992Objects}, powered by simulation \cite{battaglia2013simulation}, we propose the \longmodel~(\model) model.

We first devise a generative model that integrates physics simulation and planning (Figure~\ref{fig:InvPlan}). Given the frame of the current step, we extract the entities (the agent, objects, and obstacles) and their rough state information (3D bounding boxes and color codes), either based on the ground-truth provided in AGENT, or on results from a perception model. We then recreate an approximated physical scene in a physics engine that is different from TDW (here we use PyBullet; \citealt{coumans2019}). In particular, all obstacle entities are represented by cubes, and all objects and the agent are recreated as spheres. As the model has no access to the ground-truth parameters of the physical simulation in the procedural generation, nor any prior knowledge about the mental states of the agents, it has to propose a hypothesis of the physics parameters (coordinate transformation, global forces such as gravity and friction, and densities of entities), and a hypothesis of the agent parameters (the rewards of objects and the cost function of the agent). Given these inferred parameters, the planner (based on RRT$^*$; \citealt{karaman2011anytime}) samples a trajectory accordingly. 

We define the generative model as $ G(S^0, \Phi, \Theta)$, where $S^0 = \{s_i^0\}_{i=N}$ is the initial state of a set of entities, $N$, and $\Phi$ and $\Theta$ are the parameters for the physics engine and the agent respectively. In particular, $\Theta = (R, \mathbf{w})$, where $R = \{r_g\}_{g \in \mathcal{G}}$ indicates the agent's reward placed over a goal object $g \in \mathcal{G}$, and $C(s_a, s_a^\prime) = \mathbf{w}^\top \mathbf{f}$ is the cost function for the agent, parameterized as the weighted sum of the force needed to move the agent from its current state $s_a$ to the next state $s_a^\prime$. The generative model samples a trajectory in the next $T$ steps from $S^0$, $\hat{\Gamma}=\{s_a^t\}_{t=1}^T$, to jointly maximize the reward and minimize the cost, i.e.,
\begin{equation}
\begin{split}
\hat{\Gamma} &= G(S^0, \Phi, \Theta)\\
&=\argmax_{\Gamma=\{s_a^t\}_{t=1}^T} \sum_{g \in \mathcal{G}}r_g\delta(s_a^T, s_g) - \sum_{t=0}^{T-1} C(s_a^t, s_a^{t+1}),
\end{split}
\label{eq:gen}
\end{equation}
where $\delta(s_a^T, s_g) = 1$ if the final state of the agent ($s_a^T$) reaches goal object $g$ whose state is $s_g$, otherwise $\delta(s_a^T, s_g) = 0$. Note that we assume object-oriented goals for all agents as a built-in inductive bias. Based on Eq.~(\ref{eq:gen}), we can define the likelihood of observing an agent trajectory based on given parameters and the initial state as
\begin{equation}
P(\Gamma | S^0, \Phi, \Theta) = e^{-\beta \mathcal{D}(\Gamma, \hat{\Gamma})} = e^{-\beta \mathcal{D}(\Gamma, G(S^0, \Phi, \Theta))},
\end{equation}
where $\mathcal{D}$ is the euclidean distance between two trajectories\footnote{As two trajectories may have different lengths, we adopt dynamic time wrapping \cite{berndt1994using} for computing the distance.}, and $\beta=0.2$ adjusts the optimality of an agent's behavior. 

The training data is used to calibrate the parameters in \model. Given all $N_\text{train}$ trajectories and the corresponding initial states in the training set (from both familiarization videos and test videos), $X_\text{train} = \{(\Gamma_i, S_i^0)\}_{i\in N_\text{train}}$, we can compute the posterior probability of the parameters:
\begin{equation}
    P(\Phi, \Theta | X_\text{train}) \propto \sum_{i \in N_\text{train}} P(\Gamma_i | S_i^0, \Phi, \Theta) P(\Phi) P(\Theta)
\label{eq:post_train}
\end{equation}
where $P(\Phi)$ and $P(\Theta)$ are uniform priors of the parameters. For brevity, we define $P_\text{train}(\Phi, \Theta) = P(\Phi, \Theta | X_\text{train})$. Note that trajectories and the initial states in the videos of Unobserved Constraints are partially occluded. To obtain $X_\text{train}$, we need to reconstruct the videos. For this, we (i) first remove the occluder from the states, and (ii) reconstruct the full trajectories by applying a 2nd order curve fitting to fill the occluded the portion.

For a test trial with familiarization video(s), $X_\text{fam} = \{(\Gamma_i, S_i^0)\}_{i\in N_\text{fam}}$, and a test video, $(\Gamma_\text{test}, S^0_\text{test})$, we adjust the posterior probability of the parameters from Eq.~(\ref{eq:post_train}):

\small
\begin{equation}
    P(\Phi, \Theta | X_\text{fam}, X_\text{train}) \propto \sum_{i \in N_\text{fam}} P(\Gamma_i | S_i^0, \Phi, \Theta)P_\text{train}(\Phi, \Theta).
\label{eq:post_fam}
\end{equation}
\normalsize

We then define the surprise rating of a test video by computing the expected distance between the predicted agent trajectory and the one observed from the test video: $\E_{P(\Phi, \Theta | X_\text{fam}, X_\text{train})}\left[\mathcal{D}(\Gamma_\text{test}, G(S^0_\text{test}, \Phi, \Theta))\right]$.

\begin{figure}[t!]
\centering
\includegraphics[width=0.48\textwidth]{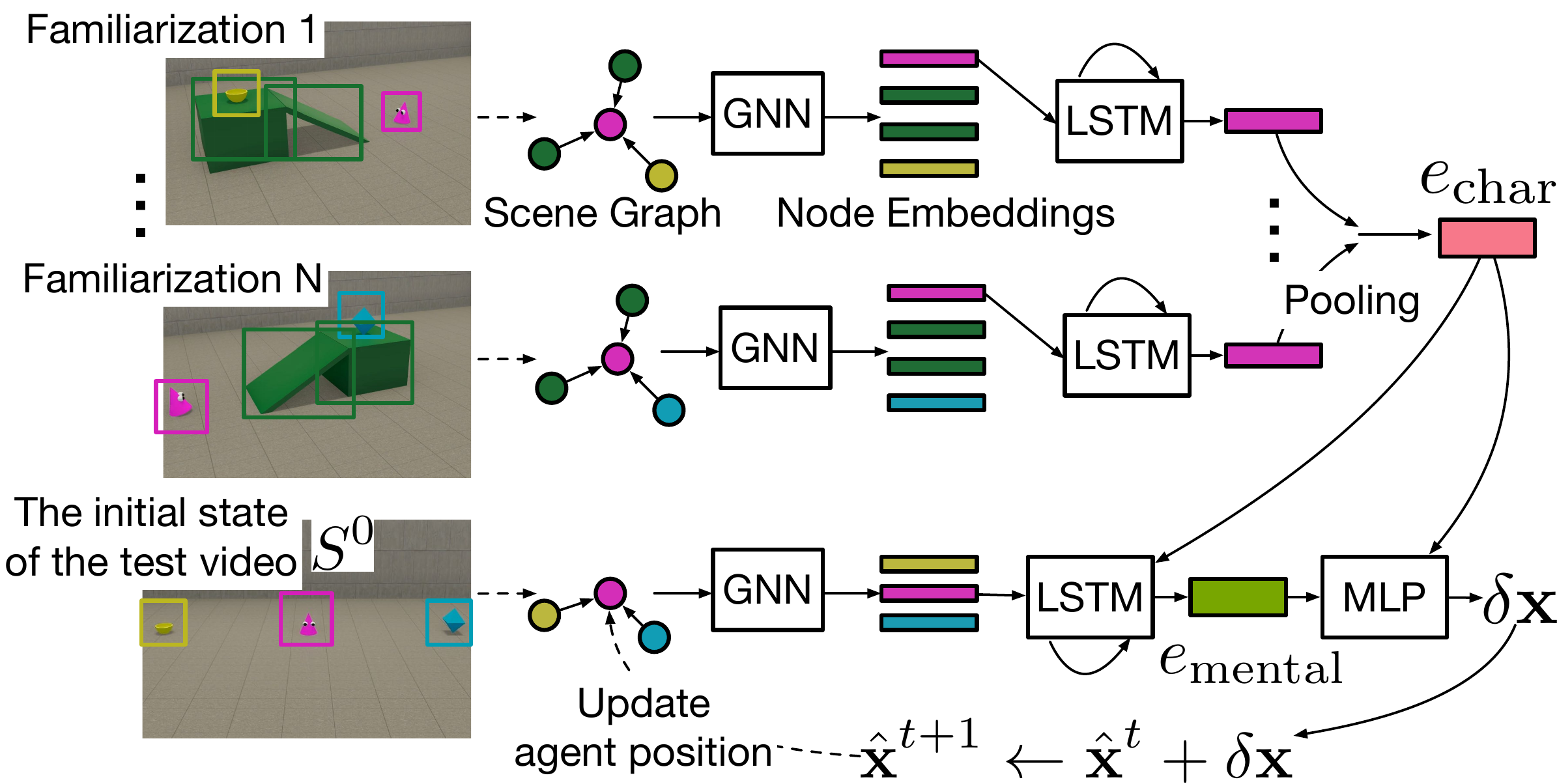}
\vspace{-15pt}
\caption{Architecture of ToMnet-G. The scene graphs are constructed based on the ground-truth or a separately trained perception model (hence the dashed arrows).}
\label{fig:ToMnet}
\end{figure}

\subsection{Theory of Mind Neural Network}\label{sec:ToMnet}

We extend ToMnet \cite{rabinowitz2018machine} to tackle the more challenging setting of AGENT, creating the second baseline model, ToMnet-G (see Figure~\ref{fig:ToMnet}). Like the original ToMnet, the network encodes the familiarization video(s) to obtain a character embedding for a particular agent, which is then combined with the embedding of the initial state to predict the expected trajectory of the agent. The surprise rating of a given test video is defined by the deviation between the predicted trajectory $\hat{\Gamma}$ and the observed trajectory $\Gamma$ in the test video. We extended ToMnet by using a graph neural network (GNN) to encode the states, where we represent all entities (including obstacles) as nodes. The input of a node includes its entity class (agent, object, obstacle), bounding box, and color code. We pass the embedding of the agent node to the downstream modules to obtain the character embedding $e_\text{char}$ and the mental state embedding $e_\text{mental}$. We train the network using a mean squared error loss on the trajectory prediction: $\mathcal{L}(\hat{\Gamma}, \Gamma) = \frac{1}{T}\sum_{i=1}^T ||\hat{\mathbf{x}}^t - \mathbf{x}^t||^2$.

To ensure that ToMnet-G can be applied to trials in Unobserved Constraints consistent with how it is applied to trials in other scenarios, we reconstruct the familiarization video and the initial state of the test video, using the same reconstruction method in Section~\ref{sec:BIP}. After the reconstruction, we can use the network to predict the expected trajectory for computing the surprise rating. Here, we use the reconstructed trajectory for calculating the surprise rating.

\begin{table*}[t!]
  \begin{center}
      \caption{Human and model performance. The `All' block reports results based on models trained on all scenarios, whereas `G1' and `G2' report model performance on `G1: leave one type out' and `G2: leave one scenario out` generalization tests. Here, G1 trains a separate model for each scenario using all but one type of trials in that scenario, and evaluates it on the held out type; G2 trains a single model on all but one scenario and evaluates it on the held out scenario. {\color{blue}Blue} numbers show where ToMnet-G generalizes well (performance $>$.8). {\color{red}Red} numbers show where it performs at or below chance (performance $\leq$.5).}
      \label{tab:main_results}
      \begin{small}
    \begin{tabular}{l|l|ccccc|cccccc|ccc|ccc|c}
    \toprule
      \parbox[t]{2mm}{\multirow{2}{*}{\rotatebox[origin=c]{90}{\textbf{Condition}}}} &\textbf{Method} & \multicolumn{5}{c|}{\textbf{Goal Preferences}} & \multicolumn{6}{c|}{\textbf{Action Efficiency}} & \multicolumn{3}{c|}{\textbf{Unobs.}} & \multicolumn{3}{c|}{\textbf{Cost-Reward}}&\textbf{All}\\
   && \multicolumn{5}{c|}{\textbf{\begin{minipage}{.08\textwidth}
      \includegraphics[width=\linewidth]{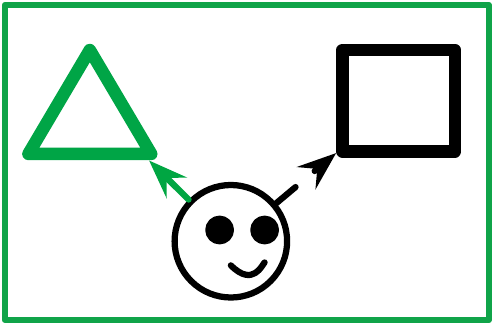}
    \end{minipage}}} & \multicolumn{6}{c|}{\begin{minipage}{.08\textwidth}
      \includegraphics[width=\linewidth]{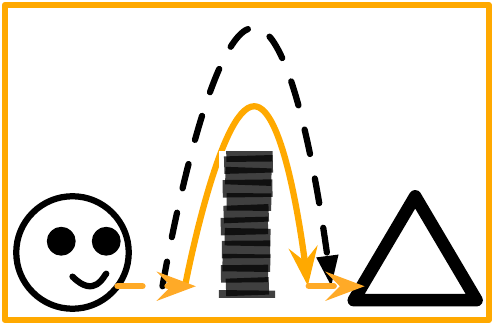}
    \end{minipage}} & \multicolumn{3}{c|}{\begin{minipage}{.08\textwidth}
      \includegraphics[width=\linewidth]{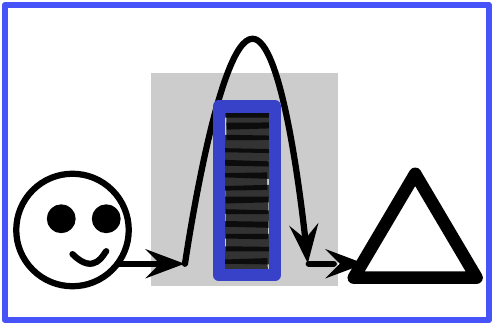}
    \end{minipage}} & \multicolumn{3}{c|}{\begin{minipage}{.08\textwidth}
      \includegraphics[width=\linewidth]{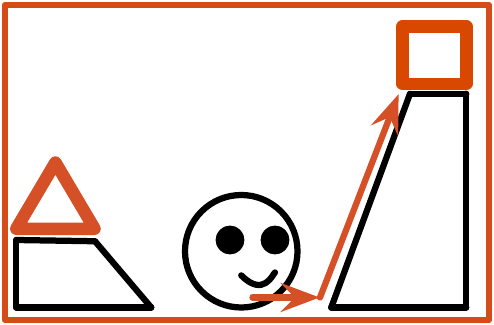}
    \end{minipage}}&\\
    
    &  &1.1&1.2&1.3 &1.4&All& 2.1&2.2&2.3&2.4&2.5&All&3.1&3.2&All&4.1&4.2&All&\\ 
      \hline
    &  Human & .95&.95&.92&.97&.95&.87&.93&.86&.95&.94&.91&.88&.94&.92&.82&
.91&.87&.91\\
      \hline
      
   \parbox[t]{1.8mm}{\multirow{2}{*}{\rotatebox[origin=c]{90}{\textbf{All}}}} &  ToMnet-G & .57 & 1.0 & .67 & 1.0 &.84 & .95 & 1.0 & .95 & 1.0 & 1.0 & .98 & .93 & .87 & .89 & .82  &	.97 & .89 & .90 \\
    &  \model~& .97 & 1.0 & 1.0 & 1.0 & .99  & 1.0 &1.0 & .85 & 1.0 & 1.0 & .97 & .93 & .88 & .90 & .90 &1.0 & .95 & .96 \\
     \hline

    \parbox[t]{1.8mm}{\multirow{2}{*}{\rotatebox[origin=c]{90}{\textbf{G1}}}} &  ToMnet-G & {\color{red}.50} & {\color{blue}.90} & .63 & {\color{blue}.88} &.75 & {\color{blue}.90} & .75 & {\color{red}.45} & {\color{blue}.90} & {\color{red}.05} & .66 & .58 & .77 & .69 & {\color{red}.48}  & {\color{red}.48} & {\color{red}.48} & .65 \\
    &  \model~& .93 & 1.0 & 1.0 & 1.0 & .98  & 1.0 &1.0 & .80 & 1.0 & 1.0 & .97 & .93 & .82 & .86 & .88 &1.0 & .94 & .94 \\
    \hline
    
    \parbox[t]{1.8mm}{\multirow{2}{*}{\rotatebox[origin=c]{90}{\textbf{G2}}}} &  ToMnet-G & {\color{red}.37} &{\color{blue}.95} & .63 & {\color{blue}.88} &.71 & {\color{red} .35} & .60 & .75 & .68 & {\color{blue}.85} & .65 & .63 & .80 & .73 & .55  &	{\color{blue}.95} & .75 & .71 \\
      
    &  \model~& .93 & 1.0 & 1.0 & 1.0 & .98  & 1.0 &1.0 & .75 & 1.0 & .95 & .95 & .88 & .85 & .87 & .83 &1.0 & .92 & .94 \\
     
      \bottomrule
    \end{tabular}
    \end{small}
  \end{center}
   \vspace{-5pt}
\end{table*}

\section{Experiments}

\subsection{Evaluation Metric} \label{sec:metric}

Following \citet{riochet2018intphys}, we define a metric based on relative surprise ratings. For a paired set of $N_{+}$ surprising test videos and $N_{-}$ expected test videos (which share the same familiarization video(s)), we obtain two sets of surprise ratings, $\{r_i^{+}\}_{i=1}^{N_{+}}$ and $\{r_j^{-}\}_{j=1}^{N_{-}}$ respectively. Accuracy is then defined as the percentage of the correctly ordered pairs of ratings: $\frac{1}{N_{+}N_{-}}\sum_{i,j} \mathds{1}(r_i^{+} > r_j^{-})$.

\subsection{Experiment 1: Human Baseline}

To validate the trials in AGENT and to estimate human baseline performance for the AGENT benchmark, we conducted an experiment in which people watched familiarization videos and then rated the relevant test videos on a sliding scale for surprise (from 0, `not at all surprising' to 100, `extremely surprising'). We randomly sampled 240 test trials (i.e., 25\% of the test set in AGENT) covering all types of trials and obstacles. We recruited 300 participants from Amazon Mechanical Turk, and each trial was rated by 10 participants. 
The participants gave informed consent, and the experiment was approved by an institutional review board.
Participants only viewed one of either the `expected' or `surprising' variants of a scene.

We found that the average human rating of each surprising video was always significantly higher than that of the corresponding expected video, resulting in a 100\% accuracy when using ratings from an ensemble of human observers. To estimate the accuracy of a single human observer, we adopted the same metric defined in Section~\ref{sec:metric}, where we first standardized the ratings of each participant so that they are directly comparable to the ratings from other participants. We report the human performance in Table~\ref{tab:main_results}.

\subsection{Experiment 2: Evaluation on Seen Scenarios and Types}

Table~\ref{tab:main_results} summarizes human performance and the performance of the two methods when the models are trained and tested on all types of trials within all four scenarios. Note that all results reported in the main paper are based on the ground-truth state information. We report the model performance based on the states extracted from a perception model in the supplementary material. When given ground-truth state information, \model~performs well on all types of trials, on par or even better than the human baseline. ToMnet-G also has a high overall accuracy when tested on all trial types it has seen during training, but performs less evenly across types within a scenario compared to \model, mostly due to overfitting certain patterns in some types. E.g., in Type 1.2 and 1.4, the agent always moves away from the object when it needs to overcome a high cost obstacle during the test phase, so ToMnet-G uses that cue to predict the agent's behavior, rather than reasoning about agent's costs and preferences given the familiarization videos (these are the kind of heuristics controls are designed to rule out in infant studies). The correlation between \model's accuracy and the human performance on different types is 0.55, versus a correlation of 0.06 between ToMnet-G and the human performance.

\begin{figure*}[t!]
\centering
\includegraphics[width=1\textwidth]{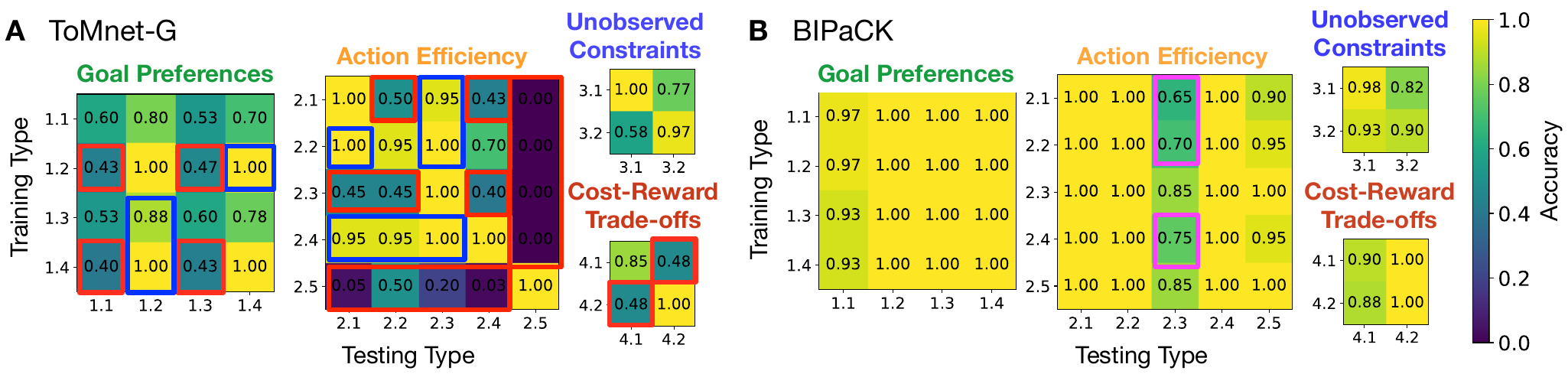}
\vspace{-10pt}
\caption{Performance of TomNet-G (A) and \model~(B) on the `G3: single type' test. This test trains a model on a single trial type within a scenario and evaluates it on the remaining types of the same scenario. {\color{blue}Blue} boxes show good generalization from ToMnet-G (off-diagonal performance $>$.8), whereas {\color{red}red} boxes show where it performs at or below chance (off-diagonal performance $\leq$.5); {\color{magenta}magenta} boxes show failures of \model~(off-diagonal performance $<$.8).}
\label{fig:generalization_one_type}
\vspace{-10pt}
\end{figure*}

\begin{figure}[t!]
\centering
\includegraphics[width=0.44\textwidth]{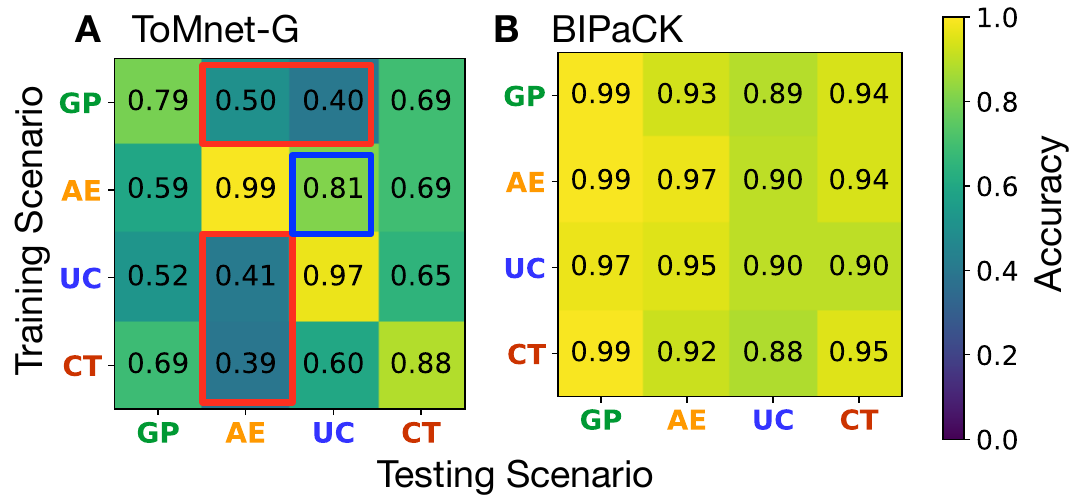}
\vspace{-5pt}
\caption{Performance of TomNet-G (A) and \model~(B) on the `G4: single scenario' test. This test trains a model on a single scenario and evaluates it on the other three scenearios. GP, AE, UC, and CT represent Goal Preferences, Action Efficiency, Unobserved Constraints, and Cost-Reward Trade-offs respectively. {\color{blue}Blue} boxes show good generalization from ToMnet-G (off-diagonal performance $>$.8, comparable to the performance when trained on the full training set), whereas {\color{red}red} boxes show where it performs at or below chance (off-diagonal performance $\leq$.5).}
\label{fig:generalization_one_concept}
\vspace{-5pt}
\end{figure}

\subsection{Experiment 3: Generalization Tests} \label{sec:exp_gen}
We conduct four types of generalization tests. The first trains a separate model for each scenario using all but one type of trials in that scenario, and evaluates it on the held out type (`G1: leave one type out'). The second trains a single model on all but one scenario and evaluates it on the held out scenario (`G2: leave one scenario out'). The third trains a model on a single trial type within a scenario and evaluates it on the remaining types of the same scenario (`G3: single type').  The fourth trains a model on a single scenario and evaluates it on the other three scenarios (`G4: single scenario'). 

We compare the performance of the two models on these four generalization tests in Table~\ref{tab:main_results} (G1 and G2), Figure~\ref{fig:generalization_one_type} (G3), and Figure~\ref{fig:generalization_one_concept} (G4). In general, we find little change in \model's performance in various generalization conditions. The largest performance drop of \model~comes from Type 2.3 (highlighted in {\color{magenta}magenta} boxes in Figure~\ref{fig:generalization_one_type}B), where the distribution of the parameters estimated from the training trials has a significant effect on the trajectory prediction (e.g., the model mistakenly predicts going around the wall, instead of the ground truth trajectory of jumping over the wall, due to an inaccurately learned cost function). In cases wherein this cost function was mis-estimated, \model~still does adjust its beliefs in the correct direction with familiarization: if it does not adjust its posterior using the familiarization video(s) (Eq.~\ref{eq:post_fam}), there would be a further 10-15\% performance drop.
ToMnet-G, on the other hand, performs well in only a few generalization conditions (e.g., results highlighted in {\color{blue}blue} in Table~\ref{tab:main_results} and in Figure~\ref{fig:generalization_one_type}A, and Figure~\ref{fig:generalization_one_concept}A). There are two main challenges that ToMnet-G faces (highlighted in {\color{red}red} in Table~\ref{tab:main_results}, Figure~\ref{fig:generalization_one_type}A, and Figure~\ref{fig:generalization_one_concept}A): (i) predicting trajectories in unfamiliar physical situations; and (ii) reliably computing costs and rewards that are grounded to objects and physics. These results complement the findings about the performance of ToMnet-based models reported in \citealt{Gandhi2021Baby}, suggesting that current model-free methods like ToMnet have a limited capacity for (i) inferring agents' mental states from a small number of familiarization videos, and (ii) generalizing the knowledge of the agents to novel situations. We report comprehensive results in the supplementary material.


\section{Conclusion}
We propose AGENT, a benchmark for core psychology reasoning, which consists of a large-scale dataset of cognitively inspired tasks designed to probe machine agents' understanding of key concepts of intuitive psychology in four scenarios -- Goal Preferences, Action Efficiency, Unobserved Constraints, and Cost-Reward Trade-offs. We validate our tasks with a large-scale set of empirical ratings from human observers, and propose several evaluation procedures that require generalization both within and across scenarios. For the proposed tasks in the benchmark, we build two baseline models (\model~and ToMnet-G) based on existing approaches, and compare their performance on AGENT to human performance. Overall, we find that \model~achieves a better performance than ToMnet-G, especially in tests of strong generalization.

Our benchmark presents exciting opportunities for future research on machine commonsense on intuitive psychology. For instance, while \model~outperforms ToMnet-G in almost all conditions, it also requires an accurate reconstruction of the 3D state and a built-in model of the physical dynamics, which will not necessarily be available in real world scenes. It is an open question whether we can learn generalizable inverse graphics and physics simulators on which BIPaCK rests. There has been work on this front (e.g., \citealt{piloto2018probing,riochet2020occlusion,wu2017learning}), from which probabilistic models built on human core knowledge of physics and psychology could potentially benefit. On the other hand, without many built-in priors, ToMnet-G demonstrates promising results when trained and tested on similar scenarios, but it still lacks a strong generalization capacity both within scenarios and across them. Generalization could be potentially improved with more advanced architectures, or pre-training on a wider variety of physical scenes to learn a more general purpose simulator.
These open areas for improvement suggest that AGENT is a well-structured diagnostic tool for developing better models of intuitive psychology. 

\section*{Acknowledgements}
This work was supported by the DARPA Machine Common Sense program, MIT-IBM AI LAB, and NSF STC award CCF-1231216.

\bibliography{references}

\begin{thebibliography}{71}
\providecommand{\natexlab}[1]{#1}
\providecommand{\url}[1]{\texttt{#1}}
\expandafter\ifx\csname urlstyle\endcsname\relax
  \providecommand{\doi}[1]{doi: #1}\else
  \providecommand{\doi}{doi: \begingroup \urlstyle{rm}\Url}\fi

\bibitem[Abbeel \& Ng(2004)Abbeel and Ng]{abbeel2004apprenticeship}
Abbeel, P. and Ng, A.~Y.
\newblock Apprenticeship learning via inverse reinforcement learning.
\newblock In \emph{Proceedings of the twenty-first international conference on
  Machine learning}, pp.\ ~1, 2004.

\bibitem[Aggarwal \& Ryoo(2011)Aggarwal and Ryoo]{aggarwal2011human}
Aggarwal, J.~K. and Ryoo, M.~S.
\newblock Human activity analysis: A review.
\newblock \emph{ACM Computing Surveys (CSUR)}, 43\penalty0 (3):\penalty0 1--43,
  2011.

\bibitem[Alahi et~al.(2016)Alahi, Goel, Ramanathan, Robicquet, Fei-Fei, and
  Savarese]{alahi2016social}
Alahi, A., Goel, K., Ramanathan, V., Robicquet, A., Fei-Fei, L., and Savarese,
  S.
\newblock Social lstm: Human trajectory prediction in crowded spaces.
\newblock In \emph{Proceedings of the IEEE conference on computer vision and
  pattern recognition}, pp.\  961--971, 2016.

\bibitem[Albrecht \& Stone(2018)Albrecht and Stone]{albrecht2018autonomous}
Albrecht, S.~V. and Stone, P.
\newblock Autonomous agents modelling other agents: A comprehensive survey and
  open problems.
\newblock \emph{Artificial Intelligence}, 258:\penalty0 66--95, 2018.

\bibitem[Baillargeon(1996)]{baillargeon1996physical}
Baillargeon, R.
\newblock Infants' understanding of the physical world.
\newblock \emph{Journal of the Neurological Sciences}, 143\penalty0
  (1-2):\penalty0 199--199, 1996.

\bibitem[Baillargeon et~al.(2016)Baillargeon, Scott, and Bian]{Baillargeon2016}
Baillargeon, R., Scott, R.~M., and Bian, L.
\newblock Psychological reasoning in infancy.
\newblock \emph{Annu. Rev. Psychol.}, 67\penalty0 (1):\penalty0 159--186, 2016.

\bibitem[Baker et~al.(2017)Baker, Jara-Ettinger, Saxe, and
  Tenenbaum]{baker2017rational}
Baker, C.~L., Jara-Ettinger, J., Saxe, R., and Tenenbaum, J.~B.
\newblock Rational quantitative attribution of beliefs, desires and percepts in
  human mentalizing.
\newblock \emph{Nature Human Behaviour}, 1\penalty0 (4):\penalty0 1--10, 2017.

\bibitem[Bakhtin et~al.(2019)Bakhtin, van~der Maaten, Johnson, Gustafson, and
  Girshick]{bakhtin2019phyre}
Bakhtin, A., van~der Maaten, L., Johnson, J., Gustafson, L., and Girshick, R.
\newblock Phyre: A new benchmark for physical reasoning.
\newblock \emph{Advances in Neural Information Processing Systems},
  32:\penalty0 5082--5093, 2019.

\bibitem[Battaglia et~al.(2013)Battaglia, Hamrick, and
  Tenenbaum]{battaglia2013simulation}
Battaglia, P.~W., Hamrick, J.~B., and Tenenbaum, J.~B.
\newblock Simulation as an engine of physical scene understanding.
\newblock \emph{Proceedings of the National Academy of Sciences}, 110\penalty0
  (45):\penalty0 18327--18332, 2013.

\bibitem[Berndt \& Clifford(1994)Berndt and Clifford]{berndt1994using}
Berndt, D.~J. and Clifford, J.
\newblock Using dynamic time warping to find patterns in time series.
\newblock In \emph{KDD workshop}, pp.\  359--370. Seattle, WA, USA:, 1994.

\bibitem[Caba~Heilbron et~al.(2015)Caba~Heilbron, Escorcia, Ghanem, and
  Carlos~Niebles]{caba2015activitynet}
Caba~Heilbron, F., Escorcia, V., Ghanem, B., and Carlos~Niebles, J.
\newblock Activitynet: A large-scale video benchmark for human activity
  understanding.
\newblock In \emph{Proceedings of the ieee conference on computer vision and
  pattern recognition}, pp.\  961--970, 2015.

\bibitem[Cao et~al.(2020)Cao, Gao, Mangalam, Cai, Vo, and Malik]{cao2020long}
Cao, Z., Gao, H., Mangalam, K., Cai, Q.-Z., Vo, M., and Malik, J.
\newblock Long-term human motion prediction with scene context.
\newblock In \emph{European Conference on Computer Vision}, pp.\  387--404.
  Springer, 2020.

\bibitem[Carroll et~al.(2019)Carroll, Shah, Ho, Griffiths, Seshia, Abbeel, and
  Dragan]{carroll2019utility}
Carroll, M., Shah, R., Ho, M.~K., Griffiths, T.~L., Seshia, S.~A., Abbeel, P.,
  and Dragan, A.
\newblock On the utility of learning about humans for human-ai coordination.
\newblock \emph{arXiv preprint arXiv:1910.05789}, 2019.

\bibitem[Choi \& Savarese(2013)Choi and Savarese]{choi2013understanding}
Choi, W. and Savarese, S.
\newblock Understanding collective activitiesof people from videos.
\newblock \emph{IEEE transactions on pattern analysis and machine
  intelligence}, 36\penalty0 (6):\penalty0 1242--1257, 2013.

\bibitem[Coumans \& Bai(2016--2019)Coumans and Bai]{coumans2019}
Coumans, E. and Bai, Y.
\newblock Pybullet, a python module for physics simulation for games, robotics
  and machine learning.
\newblock \url{http://pybullet.org}, 2016--2019.

\bibitem[Crosby et~al.(2019)Crosby, Beyret, and Halina]{crosby2019animal}
Crosby, M., Beyret, B., and Halina, M.
\newblock The animal-ai olympics.
\newblock \emph{Nature Machine Intelligence}, 1\penalty0 (5):\penalty0
  257--257, 2019.

\bibitem[Csibra et~al.(2003)Csibra, B{\'\i}r{\'o}, Ko{\'o}s, and
  Gergely]{Csibra2003Hidden}
Csibra, G., B{\'\i}r{\'o}, Z., Ko{\'o}s, O., and Gergely, G.
\newblock One-year-old infants use teleological representations of actions
  productively.
\newblock \emph{Cogn. Sci.}, 27\penalty0 (1):\penalty0 111--133, 2003.

\bibitem[Dautenhahn(2007)]{dautenhahn2007socially}
Dautenhahn, K.
\newblock Socially intelligent robots: dimensions of human--robot interaction.
\newblock \emph{Philosophical transactions of the royal society B: Biological
  sciences}, 362\penalty0 (1480):\penalty0 679--704, 2007.

\bibitem[Duan et~al.(2017)Duan, Andrychowicz, Stadie, Ho, Schneider, Sutskever,
  Abbeel, and Zaremba]{duan2017one}
Duan, Y., Andrychowicz, M., Stadie, B.~C., Ho, J., Schneider, J., Sutskever,
  I., Abbeel, P., and Zaremba, W.
\newblock One-shot imitation learning.
\newblock \emph{arXiv preprint arXiv:1703.07326}, 2017.

\bibitem[Finn et~al.(2017)Finn, Yu, Zhang, Abbeel, and Levine]{finn2017one}
Finn, C., Yu, T., Zhang, T., Abbeel, P., and Levine, S.
\newblock One-shot visual imitation learning via meta-learning.
\newblock In \emph{Conference on Robot Learning}, pp.\  357--368. PMLR, 2017.

\bibitem[Fouhey et~al.(2018)Fouhey, Kuo, Efros, and Malik]{fouhey2018lifestyle}
Fouhey, D.~F., Kuo, W.-c., Efros, A.~A., and Malik, J.
\newblock From lifestyle vlogs to everyday interactions.
\newblock In \emph{Proceedings of the IEEE Conference on Computer Vision and
  Pattern Recognition}, pp.\  4991--5000, 2018.

\bibitem[Gan et~al.(2020)Gan, Schwartz, Alter, Schrimpf, Traer, De~Freitas,
  Kubilius, Bhandwaldar, Haber, Sano, et~al.]{gan2020threedworld}
Gan, C., Schwartz, J., Alter, S., Schrimpf, M., Traer, J., De~Freitas, J.,
  Kubilius, J., Bhandwaldar, A., Haber, N., Sano, M., et~al.
\newblock Threedworld: A platform for interactive multi-modal physical
  simulation.
\newblock \emph{arXiv preprint arXiv:2007.04954}, 2020.

\bibitem[Gandhi et~al.(2021)Gandhi, Stojnic, Lake, and Dillon]{Gandhi2021Baby}
Gandhi, K., Stojnic, G., Lake, B.~M., and Dillon, M.~R.
\newblock {Baby Intuitions Benchmark (BIB): Discerning the goals, preferences,
  and actions of others}.
\newblock \emph{arXiv preprint arXiv:2102.11938}, 2021.

\bibitem[Gergely \& Csibra(2003)Gergely and Csibra]{Gergely2003}
Gergely, G. and Csibra, G.
\newblock Teleological reasoning in infancy: The na{\"\i}ve theory of rational
  action.
\newblock \emph{Trends Cogn. Sci.}, 7\penalty0 (7):\penalty0 287--292, 2003.

\bibitem[Gergely et~al.(1995)Gergely, N{\'a}dasdy, Csibra, and
  B{\'\i}r{\'o}]{Gergely1995Efficiency}
Gergely, G., N{\'a}dasdy, Z., Csibra, G., and B{\'\i}r{\'o}, S.
\newblock Taking the intentional stance at 12 months of age.
\newblock \emph{Cognition}, 56\penalty0 (2):\penalty0 165--193, 1995.

\bibitem[Gordon(2016)]{gordon2016commonsense}
Gordon, A.
\newblock Commonsense interpretation of triangle behavior.
\newblock In \emph{Proceedings of the AAAI Conference on Artificial
  Intelligence}, 2016.

\bibitem[Groth et~al.(2018)Groth, Fuchs, Posner, and
  Vedaldi]{groth2018shapestacks}
Groth, O., Fuchs, F.~B., Posner, I., and Vedaldi, A.
\newblock Shapestacks: Learning vision-based physical intuition for generalised
  object stacking.
\newblock In \emph{Proceedings of the European Conference on Computer Vision
  (ECCV)}, pp.\  702--717, 2018.

\bibitem[Hadfield-Menell et~al.(2016)Hadfield-Menell, Dragan, Abbeel, and
  Russell]{hadfield2016cooperative}
Hadfield-Menell, D., Dragan, A., Abbeel, P., and Russell, S.
\newblock Cooperative inverse reinforcement learning.
\newblock \emph{arXiv preprint arXiv:1606.03137}, 2016.

\bibitem[Huang et~al.(2019)Huang, Xu, Zhu, Garg, Savarese, Fei-Fei, and
  Niebles]{huang2019continuous}
Huang, D.-A., Xu, D., Zhu, Y., Garg, A., Savarese, S., Fei-Fei, L., and
  Niebles, J.~C.
\newblock Continuous relaxation of symbolic planner for one-shot imitation
  learning.
\newblock \emph{arXiv preprint arXiv:1908.06769}, 2019.

\bibitem[Ibrahim et~al.(2016)Ibrahim, Muralidharan, Deng, Vahdat, and
  Mori]{ibrahim2016hierarchical}
Ibrahim, M.~S., Muralidharan, S., Deng, Z., Vahdat, A., and Mori, G.
\newblock A hierarchical deep temporal model for group activity recognition.
\newblock In \emph{Proceedings of the IEEE Conference on Computer Vision and
  Pattern Recognition}, pp.\  1971--1980, 2016.

\bibitem[James et~al.(2018)James, Bloesch, and Davison]{james2018task}
James, S., Bloesch, M., and Davison, A.~J.
\newblock Task-embedded control networks for few-shot imitation learning.
\newblock In \emph{Conference on Robot Learning}, pp.\  783--795. PMLR, 2018.

\bibitem[Jara-Ettinger et~al.(2016)Jara-Ettinger, Gweon, Schulz, and
  Tenenbaum]{Jara-Ettinger2016NUC}
Jara-Ettinger, J., Gweon, H., Schulz, L.~E., and Tenenbaum, J.~B.
\newblock The na{\"\i}ve utility calculus: Computational principles underlying
  commonsense psychology.
\newblock \emph{Trends Cogn. Sci.}, 20\penalty0 (8):\penalty0 589--604, 2016.

\bibitem[Jiang et~al.(2018)Jiang, Qi, Zhu, Huang, Lin, Yu, Terzopoulos, and
  Zhu]{jiang2018configurable}
Jiang, C., Qi, S., Zhu, Y., Huang, S., Lin, J., Yu, L.-F., Terzopoulos, D., and
  Zhu, S.-C.
\newblock Configurable 3d scene synthesis and 2d image rendering with per-pixel
  ground truth using stochastic grammars.
\newblock \emph{International Journal of Computer Vision}, 126\penalty0
  (9):\penalty0 920--941, 2018.

\bibitem[Johnson et~al.(2017)Johnson, Hariharan, van~der Maaten, Fei-Fei,
  Lawrence~Zitnick, and Girshick]{johnson2017clevr}
Johnson, J., Hariharan, B., van~der Maaten, L., Fei-Fei, L., Lawrence~Zitnick,
  C., and Girshick, R.
\newblock Clevr: A diagnostic dataset for compositional language and elementary
  visual reasoning.
\newblock In \emph{Proceedings of the IEEE Conference on Computer Vision and
  Pattern Recognition}, pp.\  2901--2910, 2017.

\bibitem[Karaman et~al.(2011)Karaman, Walter, Perez, Frazzoli, and
  Teller]{karaman2011anytime}
Karaman, S., Walter, M.~R., Perez, A., Frazzoli, E., and Teller, S.
\newblock Anytime motion planning using the rrt$^*$.
\newblock In \emph{2011 IEEE International Conference on Robotics and
  Automation}, pp.\  1478--1483. IEEE, 2011.

\bibitem[Kitani et~al.(2012)Kitani, Ziebart, Bagnell, and
  Hebert]{kitani2012activity}
Kitani, K.~M., Ziebart, B.~D., Bagnell, J.~A., and Hebert, M.
\newblock Activity forecasting.
\newblock In \emph{European Conference on Computer Vision}, pp.\  201--214.
  Springer, 2012.

\bibitem[Kleiman-Weiner et~al.(2016)Kleiman-Weiner, Ho, Austerweil, Littman,
  and Tenenbaum]{kleiman2016coordinate}
Kleiman-Weiner, M., Ho, M.~K., Austerweil, J.~L., Littman, M.~L., and
  Tenenbaum, J.~B.
\newblock Coordinate to cooperate or compete: abstract goals and joint
  intentions in social interaction.
\newblock In \emph{CogSci}, 2016.

\bibitem[Kong \& Fu(2018)Kong and Fu]{kong2018human}
Kong, Y. and Fu, Y.
\newblock Human action recognition and prediction: A survey.
\newblock \emph{arXiv preprint arXiv:1806.11230}, 2018.

\bibitem[Koppula \& Saxena(2013)Koppula and Saxena]{koppula2013learning}
Koppula, H. and Saxena, A.
\newblock Learning spatio-temporal structure from rgb-d videos for human
  activity detection and anticipation.
\newblock In \emph{International conference on machine learning}, pp.\
  792--800. PMLR, 2013.

\bibitem[Liang et~al.(2019)Liang, Jiang, Niebles, Hauptmann, and
  Fei-Fei]{liang2019peeking}
Liang, J., Jiang, L., Niebles, J.~C., Hauptmann, A.~G., and Fei-Fei, L.
\newblock Peeking into the future: Predicting future person activities and
  locations in videos.
\newblock In \emph{Proceedings of the IEEE/CVF Conference on Computer Vision
  and Pattern Recognition}, pp.\  5725--5734, 2019.

\bibitem[Liu et~al.(2017)Liu, Ullman, Tenenbaum, and Spelke]{Liu2017}
Liu, S., Ullman, T.~D., Tenenbaum, J.~B., and Spelke, E.~S.
\newblock Ten-month-old infants infer the value of goals from the costs of
  actions.
\newblock \emph{Science}, 358\penalty0 (6366):\penalty0 1038--1041, November
  2017.

\bibitem[Lowe et~al.(2017)Lowe, Wu, Tamar, Harb, Abbeel, and
  Mordatch]{lowe2017multi}
Lowe, R., Wu, Y., Tamar, A., Harb, J., Abbeel, P., and Mordatch, I.
\newblock Multi-agent actor-critic for mixed cooperative-competitive
  environments.
\newblock \emph{arXiv preprint arXiv:1706.02275}, 2017.

\bibitem[Mordatch \& Abbeel(2018)Mordatch and Abbeel]{mordatch2018emergence}
Mordatch, I. and Abbeel, P.
\newblock Emergence of grounded compositional language in multi-agent
  populations.
\newblock In \emph{Proceedings of the AAAI Conference on Artificial
  Intelligence}, 2018.

\bibitem[Nan et~al.(2020)Nan, Shu, Gong, Wang, Wei, Zhu, and
  Zheng]{nan2020learning}
Nan, Z., Shu, T., Gong, R., Wang, S., Wei, P., Zhu, S.-C., and Zheng, N.
\newblock Learning to infer human attention in daily activities.
\newblock \emph{Pattern Recognition}, pp.\  107314, 2020.

\bibitem[Netanyahu et~al.(2021)Netanyahu, Shu, Katz, Barbu, and
  Tenenbaum]{phase}
Netanyahu, A., Shu, T., Katz, B., Barbu, A., and Tenenbaum, J.~B.
\newblock {PHASE: PHysically-grounded Abstract Social Events for machine social
  perception}.
\newblock In \emph{Proceedings of the AAAI Conference on Artificial
  Intelligence (AAAI)}, 2021.

\bibitem[Nikolaidis et~al.(2017)Nikolaidis, Hsu, and
  Srinivasa]{nikolaidis2017human}
Nikolaidis, S., Hsu, D., and Srinivasa, S.
\newblock Human-robot mutual adaptation in collaborative tasks: Models and
  experiments.
\newblock \emph{The International Journal of Robotics Research}, 36\penalty0
  (5-7):\penalty0 618--634, 2017.

\bibitem[Piloto et~al.(2018)Piloto, Weinstein, TB, Ahuja, Mirza, Wayne, Amos,
  Hung, and Botvinick]{piloto2018probing}
Piloto, L., Weinstein, A., TB, D., Ahuja, A., Mirza, M., Wayne, G., Amos, D.,
  Hung, C.-c., and Botvinick, M.
\newblock Probing {{Physics Knowledge Using Tools}} from {{Developmental
  Psychology}}.
\newblock \emph{arXiv:1804.01128 [cs]}, 2018.

\bibitem[Poppe(2010)]{poppe2010survey}
Poppe, R.
\newblock A survey on vision-based human action recognition.
\newblock \emph{Image and vision computing}, 28\penalty0 (6):\penalty0
  976--990, 2010.

\bibitem[Puig et~al.(2020)Puig, Shu, Li, Wang, Tenenbaum, Fidler, and
  Torralba]{puig2020watch}
Puig, X., Shu, T., Li, S., Wang, Z., Tenenbaum, J.~B., Fidler, S., and
  Torralba, A.
\newblock {Watch-And-Help: A Challenge for Social Perception and Human-AI
  Collaboration}.
\newblock \emph{arXiv preprint arXiv:2010.09890}, 2020.

\bibitem[Rabinowitz et~al.(2018)Rabinowitz, Perbet, Song, Zhang, Eslami, and
  Botvinick]{rabinowitz2018machine}
Rabinowitz, N., Perbet, F., Song, F., Zhang, C., Eslami, S.~A., and Botvinick,
  M.
\newblock Machine theory of mind.
\newblock In \emph{International conference on machine learning}, pp.\
  4218--4227. PMLR, 2018.

\bibitem[Riochet et~al.(2018)Riochet, Castro, Bernard, Lerer, Fergus, Izard,
  and Dupoux]{riochet2018intphys}
Riochet, R., Castro, M.~Y., Bernard, M., Lerer, A., Fergus, R., Izard, V., and
  Dupoux, E.
\newblock {{IntPhys}}: {{A Framework}} and {{Benchmark}} for {{Visual Intuitive
  Physics Reasoning}}.
\newblock \emph{arXiv:1803.07616 [cs]}, 2018.

\bibitem[Riochet et~al.(2020)Riochet, Sivic, Laptev, and
  Dupoux]{riochet2020occlusion}
Riochet, R., Sivic, J., Laptev, I., and Dupoux, E.
\newblock Occlusion resistant learning of intuitive physics from videos.
\newblock \emph{arXiv:2005.00069 [cs, eess]}, 2020.

\bibitem[Ros et~al.(2016)Ros, Sellart, Materzynska, Vazquez, and
  Lopez]{ros2016synthia}
Ros, G., Sellart, L., Materzynska, J., Vazquez, D., and Lopez, A.~M.
\newblock The synthia dataset: A large collection of synthetic images for
  semantic segmentation of urban scenes.
\newblock In \emph{Proceedings of the IEEE conference on computer vision and
  pattern recognition}, pp.\  3234--3243, 2016.

\bibitem[Rozo et~al.(2016)Rozo, Calinon, Caldwell, Jimenez, and
  Torras]{rozo2016learning}
Rozo, L., Calinon, S., Caldwell, D.~G., Jimenez, P., and Torras, C.
\newblock Learning physical collaborative robot behaviors from human
  demonstrations.
\newblock \emph{IEEE Transactions on Robotics}, 32\penalty0 (3):\penalty0
  513--527, 2016.

\bibitem[Sadigh et~al.(2016)Sadigh, Sastry, Seshia, and
  Dragan]{sadigh2016planning}
Sadigh, D., Sastry, S., Seshia, S.~A., and Dragan, A.~D.
\newblock Planning for autonomous cars that leverage effects on human actions.
\newblock In \emph{Robotics: Science and Systems}, volume~2. Ann Arbor, MI,
  USA, 2016.

\bibitem[Sheridan(2016)]{sheridan2016human}
Sheridan, T.~B.
\newblock Human--robot interaction: status and challenges.
\newblock \emph{Human factors}, 58\penalty0 (4):\penalty0 525--532, 2016.

\bibitem[Shu \& Tian(2018)Shu and Tian]{shu2018m}
Shu, T. and Tian, Y.
\newblock {M$^3$RL: Mind-aware Multi-agent Management Reinforcement Learning}.
\newblock \emph{arXiv preprint arXiv:1810.00147}, 2018.

\bibitem[Shu et~al.(2015)Shu, Xie, Rothrock, Todorovic, and Zhu]{shu2015joint}
Shu, T., Xie, D., Rothrock, B., Todorovic, S., and Zhu, S.-C.
\newblock Joint inference of groups, events and human roles in aerial videos.
\newblock In \emph{Proceedings of the IEEE Conference on Computer Vision and
  Pattern Recognition}, pp.\  4576--4584, 2015.

\bibitem[Sigurdsson et~al.(2018)Sigurdsson, Gupta, Schmid, Farhadi, and
  Alahari]{sigurdsson2018charades}
Sigurdsson, G.~A., Gupta, A., Schmid, C., Farhadi, A., and Alahari, K.
\newblock Charades-ego: A large-scale dataset of paired third and first person
  videos.
\newblock \emph{arXiv preprint arXiv:1804.09626}, 2018.

\bibitem[Silver et~al.(2020)Silver, Allen, Lew, Kaelbling, and
  Tenenbaum]{silver2020few}
Silver, T., Allen, K.~R., Lew, A.~K., Kaelbling, L.~P., and Tenenbaum, J.
\newblock Few-shot bayesian imitation learning with logical program policies.
\newblock In \emph{Proceedings of the AAAI Conference on Artificial
  Intelligence}, pp.\  10251--10258, 2020.

\bibitem[Song et~al.(2017)Song, Yu, Zeng, Chang, Savva, and
  Funkhouser]{song2017semantic}
Song, S., Yu, F., Zeng, A., Chang, A.~X., Savva, M., and Funkhouser, T.
\newblock Semantic scene completion from a single depth image.
\newblock In \emph{Proceedings of the IEEE Conference on Computer Vision and
  Pattern Recognition}, pp.\  1746--1754, 2017.

\bibitem[Spelke et~al.(1992)Spelke, Breinlinger, Macomber, and
  Jacobson]{Spelke1992Objects}
Spelke, E.~S., Breinlinger, K., Macomber, J., and Jacobson, K.
\newblock Origins of knowledge.
\newblock \emph{Psychol. Rev.}, 99\penalty0 (4):\penalty0 605--632, October
  1992.

\bibitem[Wang et~al.(2020)Wang, Wu, Evans, Tenenbaum, Parkes, and
  Kleiman-Weiner]{wang2020too}
Wang, R.~E., Wu, S.~A., Evans, J.~A., Tenenbaum, J.~B., Parkes, D.~C., and
  Kleiman-Weiner, M.
\newblock Too many cooks: Bayesian inference for coordinating multi-agent
  collaboration.
\newblock \emph{arXiv e-prints}, pp.\  arXiv--2003, 2020.

\bibitem[Woodward(1998)]{Woodward1998}
Woodward, A.~L.
\newblock Infants selectively encode the goal object of an actor's reach.
\newblock \emph{Cognition}, 69\penalty0 (1):\penalty0 1--34, 1998.

\bibitem[Wu et~al.(2017)Wu, Lu, Kohli, Freeman, and Tenenbaum]{wu2017learning}
Wu, J., Lu, E., Kohli, P., Freeman, W.~T., and Tenenbaum, J.~B.
\newblock Learning to {{See Physics}} via {{Visual De}}-animation.
\newblock In \emph{Neural {{Information Processing Systems}}}, pp.\ ~12, 2017.

\bibitem[Xia et~al.(2018)Xia, R.~Zamir, He, Sax, Malik, and
  Savarese]{xiazamirhe2018gibsonenv}
Xia, F., R.~Zamir, A., He, Z.-Y., Sax, A., Malik, J., and Savarese, S.
\newblock Gibson env: real-world perception for embodied agents.
\newblock In \emph{Computer Vision and Pattern Recognition (CVPR), 2018 IEEE
  Conference on}. IEEE, 2018.

\bibitem[Xie et~al.(2020)Xie, Losey, Tolsma, Finn, and Sadigh]{xie2020learning}
Xie, A., Losey, D.~P., Tolsma, R., Finn, C., and Sadigh, D.
\newblock Learning latent representations to influence multi-agent interaction.
\newblock \emph{arXiv preprint arXiv:2011.06619}, 2020.

\bibitem[Yi et~al.(2019)Yi, Gan, Li, Kohli, Wu, Torralba, and
  Tenenbaum]{yi2019clevrer}
Yi, K., Gan, C., Li, Y., Kohli, P., Wu, J., Torralba, A., and Tenenbaum, J.~B.
\newblock Clevrer: Collision events for video representation and reasoning.
\newblock \emph{arXiv preprint arXiv:1910.01442}, 2019.

\bibitem[Yu et~al.(2018)Yu, Finn, Xie, Dasari, Zhang, Abbeel, and
  Levine]{yu2018one}
Yu, T., Finn, C., Xie, A., Dasari, S., Zhang, T., Abbeel, P., and Levine, S.
\newblock One-shot imitation from observing humans via domain-adaptive
  meta-learning.
\newblock \emph{arXiv preprint arXiv:1802.01557}, 2018.

\bibitem[Ziebart et~al.(2008)Ziebart, Maas, Bagnell, and
  Dey]{ziebart2008maximum}
Ziebart, B.~D., Maas, A.~L., Bagnell, J.~A., and Dey, A.~K.
\newblock Maximum entropy inverse reinforcement learning.
\newblock In \emph{Aaai}, volume~8, pp.\  1433--1438. Chicago, IL, USA, 2008.

\bibitem[Zitnick et~al.(2014)Zitnick, Vedantam, and
  Parikh]{zitnick2014adopting}
Zitnick, C.~L., Vedantam, R., and Parikh, D.
\newblock Adopting abstract images for semantic scene understanding.
\newblock \emph{IEEE transactions on pattern analysis and machine
  intelligence}, 38\penalty0 (4):\penalty0 627--638, 2014.

\end{thebibliography}
\bibliographystyle{icml2021}

\end{document}